\documentclass[accepted]{uai2023} 

\usepackage[american]{babel}

\usepackage{natbib} 
\usepackage{mathtools} 
\usepackage{booktabs} 
\usepackage{tikz} 
\usepackage{multirow}
\usepackage{tabularx}
\usepackage{algorithm}
\usepackage{algorithmic}

\newcommand\x{\mathbf{x}}
\newcommand\bC{\mathbf{C}}
\newcommand\bx{\mathbf{x}}

\title{Two-Stage Holistic and Contrastive Explanation of Image Classification}

\author[1]{\href{mailto:<wxieai@cse.ust.hk>}{Weiyan Xie}\ $^*$}
\author[2]{\href{mailto:<lixiaohui33@huawei.com>}{Xiao-Hui Li}\ }
\author[1]{\href{mailto:<zlinaz@cse.ust.hk>}{Zhi Lin}\ }
\author[3]{\href{mailto:<leonard.poon@gmail.com>}{Leonard K. M. Poon}\ }
\author[1]{\href{mailto:<cao@ust.hk>}{Caleb Chen Cao}\ $^\dag$} 
\author[1]{\href{mailto:<lzhang@cse.ust.hk>}{Nevin L. Zhang}\ $^*$}
\affil[1]{
    The Hong Kong University of Science and Technology\\
    Hong Kong, China
}
\affil[2]{
   Huawei Technologies Co., Ltd\\ Shenzhen, China
}
\affil[3]{
  The Education University of Hong Kong\\
  Hong Kong, China$^\ddag$
  }

\DeclareMathOperator\supp{supp}

\begin{document}
\maketitle

\def\thefootnote{*}\footnotetext{Equal Contribution.}\def\thefootnote{\arabic{footnote}}
\def\thefootnote{$\dag$}\footnotetext{This work is done when Caleb was in Huawei Research HK.}\def\thefootnote{\arabic{footnote}}
\def\thefootnote{$\ddag$}\footnotetext{Previous affiliation where the work was carried out.}\def\thefootnote{\arabic{footnote}}

\begin{abstract}
The need to explain the output of a deep neural network classifier is now widely recognized. While previous methods typically explain a single class in the output, we advocate explaining the whole output, which is a probability distribution over multiple classes. A whole-output explanation can help a human user gain an overall	understanding of model behaviour instead of only one aspect of it. It can also provide a natural framework where one can examine the evidence used to discriminate between	competing classes, and thereby obtain contrastive explanations. In this paper, we propose a contrastive whole-output explanation (CWOX) method for image classification, and evaluate it using quantitative metrics and through  human subject studies. The source code of CWOX is available at \url{https://github.com/vaynexie/CWOX}.
\end{abstract}

\section{Introduction}
\label{sec.intro}

The past few years have witnessed a surge of research activities on the explainability of deep neural networks, which is driven by the need for trust,  fairness and accountability in high-stake applications \citep{samek2019explainable,li2020survey}.
While there is some work on {\em ante hoc methods} that learn interpretable models to begin with \citep{zhang2018interpretable}, most efforts are spent on {\em post hoc methods} that explain complex models whose behaviours are not self-interpretable   \citep{samek2019explainable,li2020survey}.
A common way to explain image classification  is to generate a saliency map that assigns a numerical value to each pixel to indicate its importance to an output class label. A variety of methods have been proposed
\citep{simonyan2013deep,springenberg2014striving,zeiler2014visualizing, bach2015pixel,ribeiro2016should,shrikumar2017learning,zhang2018top,petsiuk2018rise}. Most methods are designed to explain one single output label, and hence we call them {\em individual output explanation (IOX)} methods.

\begin{figure}[t]
	\centering
	\includegraphics[width=8.2cm]{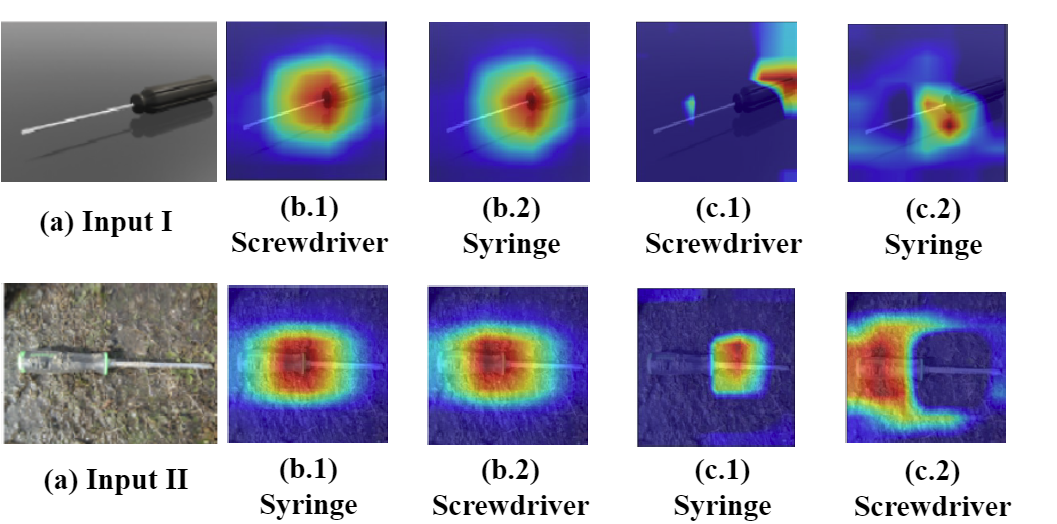}\\
	\caption{Explanations of the outputs of GoogleNet on two input images: (b.1-2) Grad-CAM is applied to each top output class separately (SWOX); (c.1-2) The top output classes are contrasted against each other (CWOX).}
	\label{fig.syrdinge.intro}
\end{figure}

IOX methods are unable to provide users with an overall understanding of model behavior \citep{kim2021machine}, and might mislead users to unjustified confidence in the explanation and the  model \citep{rudin2019stop,adebayo2018sanity,adebayo2022post}. Consider the two input images in  Fig.\ \ref{fig.syrdinge.intro}, both with ground-truth label {\tt screwdriver}.  The outputs of GoogleNet \citep{szegedy2015going} are 
\{{\tt screwdriver} (0.49), {\tt syringe} (0.38)\} ({\em input I}), and  \{{\tt syringe} (0.50), {\tt screwdriver} (0.38)\} ({\em input II}) respectively.   The saliency maps created using Grad-CAM \citep{shrikumar2017learning} for the output classes are shown in (b.1-2).

Consider two scenarios for the first input in Fig.\ \ref{fig.syrdinge.intro}\ : (1) Present a user only with the 
heatmap for the top class (b.1), or (2) present a user with the heatmaps for both top classes (b.1-2). Clearly, the user would gain a better understanding of the model in the second scenario and realize that the model has difficulty in discriminating {\tt screwdriver} and {\tt syringe}. In addition, the user would realize that the two heatmaps, being almost identical, do not help  understand what evidence the model uses to discriminate the two classes. To appreciate the point better, imagine a scenario where a user is  presented with the two heatmaps and the two labels {\em separately}, and is asked to match them. This would be virtually an impossible task. The same is true for the second input, where the order of the top 2 labels is reversed.

It is clear  that we need
{\em whole-output explanation (WOX)}  methods that 
explain all top output classes. It is also evident that 
a {\em simple WOX (SWOX)} method, which explains the top classes one by one independently, is not sufficient.  
It is necessary to reveal the evidence that supports each top class against other top classes~\citep{wang2020scout}.  This leads to what we call {\em Contrastive Whole-Output Explanation (CWOX)}.  For the first example in Fig.\ \ref{fig.syrdinge.intro}, the CWOX explanations are shown in top row (c.1-2).  We see that the handle is highlighted for {\tt screwdriver} and the shaft is highlighted for {\tt syringe}.  Those can evidently help a user understand why there are two possible output classes instead of one, and correctly match the heatmaps with the label in the case where there are presented separately. The same is true for the second example.

\begin{figure}[t]
	\centering
	\includegraphics[width=8.27cm]{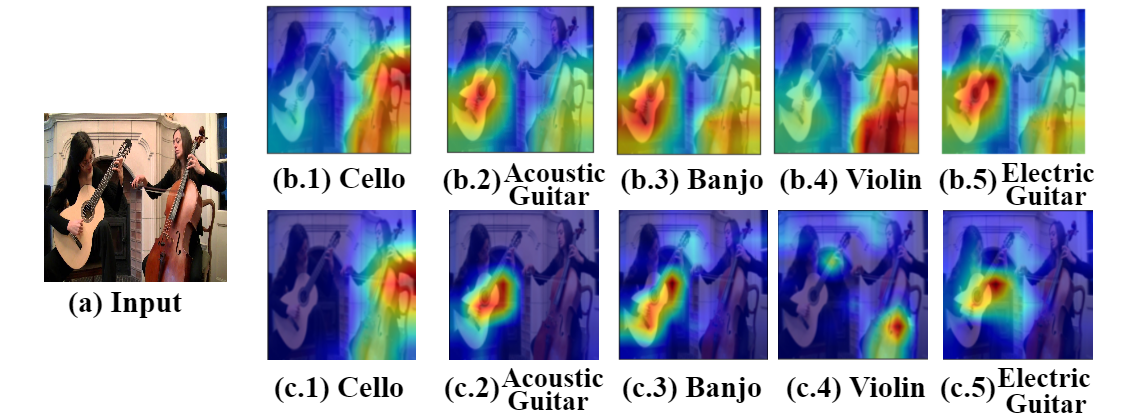}\\
	\caption{The output of ResNet50 on the input image includes
	5 top classes. The top row (b.1-5) shows their SWOX saliency maps while the bottom row (c.1-5) show their CWOX-1s saliency maps.}
	\label{fig.cello.intro}
\end{figure}

Images often contain multiple objects of interest.  
Compared with those with a single object,
such images usually lead to more classes with significant probabilities in model output.  For example, the output of  ResNet50 \citep{he2016deep} on
the input image shown in Fig.\ \ref{fig.cello.intro} consists of 5 top classes: {\tt cello} (0.839), {\tt acoustic-guitar} (0.081), {\tt banjo} (0.036), {\tt violin} (0.021), {\tt electric-guitar} (0.008).
From the SWOX saliency maps (top row in Fig.\ \ref{fig.cello.intro}),  
 we see that different top classes (e.g., {\tt cello} and {\tt violin}) might refer to the same object in the input image and are competing labels for that object.  Such classes are {\em confusing} to the classifier in the sense that the classifier is uncertain as to which of the classes to use when labeling the object.

\begin{figure}[t]
	\centering
	\includegraphics[width=8.15cm]{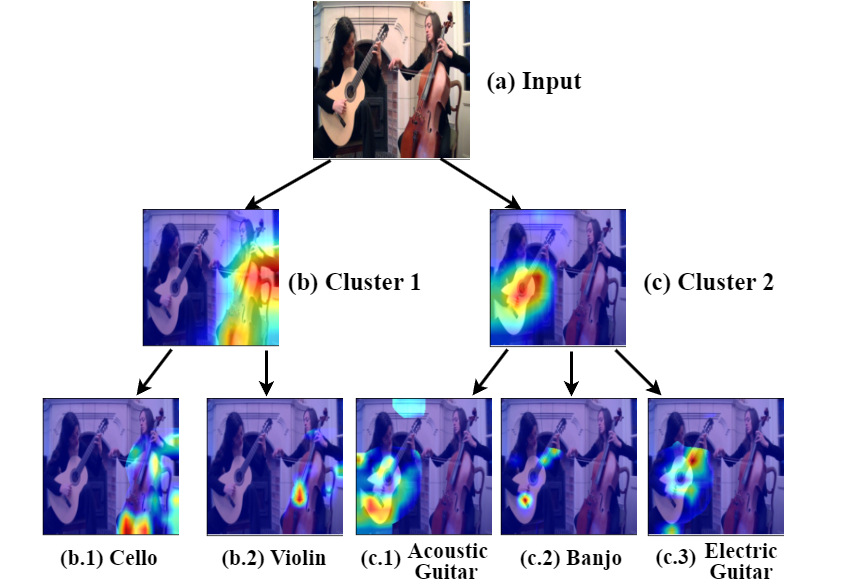}\\
	\caption{CWOX-2s explanation of the output of ResNet on the input image (with Grad-CAM as the base explainer).}
	\label{c-wox}
\end{figure}

Our main contribution in this paper is to show that the quality of explanations can be substantially improved by utilizing this observation.  Specifically,  we propose to divide the top class labels into confusion clusters based on the object they refer to, and perform the explanation in two steps: (1) Generate heatmaps to contrast different confusion clusters, and (2) generate heatmaps to contrast classes within each cluster.  We call this method {\em two-stage contrastive whole-output explanation (CWOX-2s)}.  On the other hand, the method  alluded to in previous paragraphs contrasts each class directly against all other classes.  We call it {\em one-stage contrastive whole-output explanation (CWOX-1s)}. Note that CWOX-2s reduces to CWOX-1s when there is only one confusion cluster, as in the case of Fig.\ \ref{fig.syrdinge.intro}. However, CWOX-2s makes significantly different explanations  when there are more than one confusion clusters.
 
For instance, in the example shown in Fig.~\ref{fig.cello.intro}, CWOX-2s divides the top five classes into two clusters: \{{\tt cello}, {\tt violin}\} and \{{\tt acoustic guitar}, {\tt banjo}, {\tt electric guitar}\}.  It first contrasts the two clusters, and then contrasts classes within each cluster against the other classes in the same cluster. This approach is more reasonable than CWOX-1s. In Fig.~\ref{fig.cello.intro}, it is clear that {\tt violin}  should have more contrastive value to {\tt cello} than other classes. This observation is ignored by CWOX-1s.

The explanation given by CWOX-2s is as shown in
Fig.~\ref{c-wox}. It first shows that evidence for the two clusters comes from the left and right part of the input image, respectively. {\tt Cello}  and {\tt violin} are competing labels for the right part of the image.  The evidence
that supports {\tt cello}  relative to {\tt violin} is the body bottom of the instrument (b.1), and the evidence that supports {\tt violin}  relative to {\tt cello} is the middle section of the strings (b.2). Those make sense intuitively because cellos have large bottoms and the middle section of the strings on a cello is visually similar to that on a violin. The supportive evidence for the three labels in the other cluster relative to each other are displayed in (c.1) (lower body), (c.2) (bridge), and (c.3) (strings), respectively.   Those are intuitively more informative than the heatmaps by CWOX-1s shown in the second row of Fig.\ \ref{fig.cello.intro}. Later we will show that CWOX-2s is superior to SWOX and CWOX-1s in both quantitative evaluations and human subject studies.

\section{Related Work}
\label{sec.related}

CWOX-2s aims to provide contrastive explanations for the top predicted classes. There are previous works on contrastive explanations. \citet{miller2019explanation} surveyed over 250 papers in philosophy, psychology, and cognitive science and found that humans prefer contrastive explanations that explain {\em why class $A$ but not class $B$} to non-contrastive ones that only explain {\em why $A$}. In XAI, this is achieved  through {\em counterfactual explanation} or {\em discriminative explanation}. Counterfactual explanations identify necessary modifications to change the prediction from $A$ to $B$  \citep{wachter2017counterfactual}, while discriminative explanations provide the evidence in the input that supports $A$ over $B$ \citep{wang2020scout,prabhushankar2020contrastive,jacovi2021contrastive}. As illustrated in Fig.\ \ref{c-wox}, CWOX-2s is a systematic and organized way to apply discriminative explanation to the top classes in the classification output.

In both types of contrastive explanations,  there is a need to identify a {\em contrast class (foil)}  $B$  for the {\em target class (fact) $A$}. Previous works  let the foil be: (1) all other classes (i.e., non-$A$) \citep{zhang2018top,jacovi2021contrastive};  (2) any other class  \citep{dhurandhar2018explanations,goyal2019counterfactual,wang2020scout}; (3) the class with second highest probability \citep{wang2022why}; (4) another class picked by users \citep{liu2019generative,akula2020cocox}; or  (5) the prediction of another smaller model \citep{wang2020scout}. In CWOX-2s, we propose a principled method for determining how to contrast the top classes against each other. Specifically, we divide the top classes into confusion clusters. We first contrast different confusion clusters against each other, and then contrast different classes within the same confusion cluster. 

In XAI literature, methods explaining ``{\em why class $A$}" and methods explaining ``{\em why class $A$ but not class $B$}" are regarded as two separate lines of work. The first line of work is essentially about localizing the object (or region) that class $A$ refers to \citep{selvaraju2017grad,shrikumar2017learning,zhang2018top}. The second line of work is about deciding whether the object in focus belongs to class $A$ or $B$ \citep{dhurandhar2018explanations,prabhushankar2020contrastive,goyal2019counterfactual,wang2020scout}. The latter is carried out in the context of fine-grained image classification. CWOX-2s can be viewed as a junction where the two lines of work meet. The first step of CWOX-2s is about object localization, and the second step targets class discrimination.

\section{Grouping Class Labels into Confusion Clusters}
\label{sec.hlta}

As mentioned in Section \ref{sec.intro}, the notion of confusion clusters is of pivotal importance to CWOX-2s.   The question arises: how do we divide the top output labels for a given input into confusion clusters? A straightforward approach is to examine the IOX heatmaps for all the top labels and group two labels together into the same cluster if their IOX heatmaps overlap substantially. A threshold is required for this approach. We have found it difficult to determine a threshold that suits all cases.  Consequently, we step back and ask how to tell if two classes are confusing to a classifier without using XAI?  One answer is described below. It is one of the main innovative aspects of this paper.

\begin{figure}[t]
	\centering
	\includegraphics[width=7.5cm]{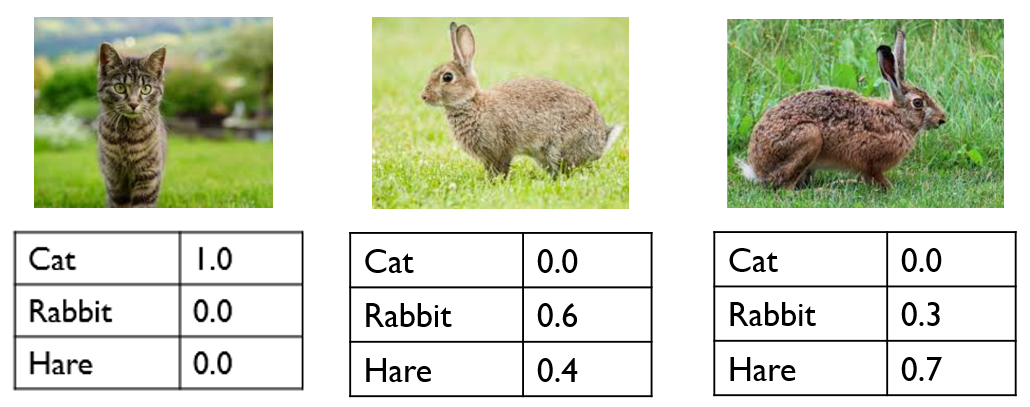}
	\caption{Rabbit and hare are confusing to humans.  A
		person would find it difficult to decide whether to label
		the second and third images as rabbit or hare. Similarly, a neural network classifier would give the two class labels high probabilities in either case.
	}
	\label{rabbit-hare}
\end{figure}

Rabbit and hare are confusing to humans. When presented with an image of either  class, a person would find it difficult to decide whether to label it as a rabbit or a hare (Fig.~\ref{rabbit-hare}).  Similarly, two classes are confusing to a classifier if,  when processing images containing objects of either class, it has trouble determining which of them to use as the output label. Consequently,  it gives  high probabilities to both classes.
{It is therefore possible to determine if two classes are confusing to a classifier by checking if they {\em often} co-occur as top classes in classification outputs}.

To partition classes into confusion clusters, we first run the target classifier  on a set of examples, typically the training examples. For each example, we get a list of top class labels, which we regard as a short document. For the example in Fig.~\ref{c-wox}, the document consists of five words: \{{\tt cello}, {\tt acoustic-guitar}, {\tt banjo}, {\tt violin}, {\tt electric-guitar}\}.
Suppose there are $N$ training examples.  Then we have $N$ short documents.
The task now becomes a word clustering problem. We want to partition the words (class labels) into clusters such that words  from the same cluster co-occur more often in the $N$ documents than words from different clusters. 

\begin{figure*}[t]
	\centering
	\includegraphics[width=12.5cm]{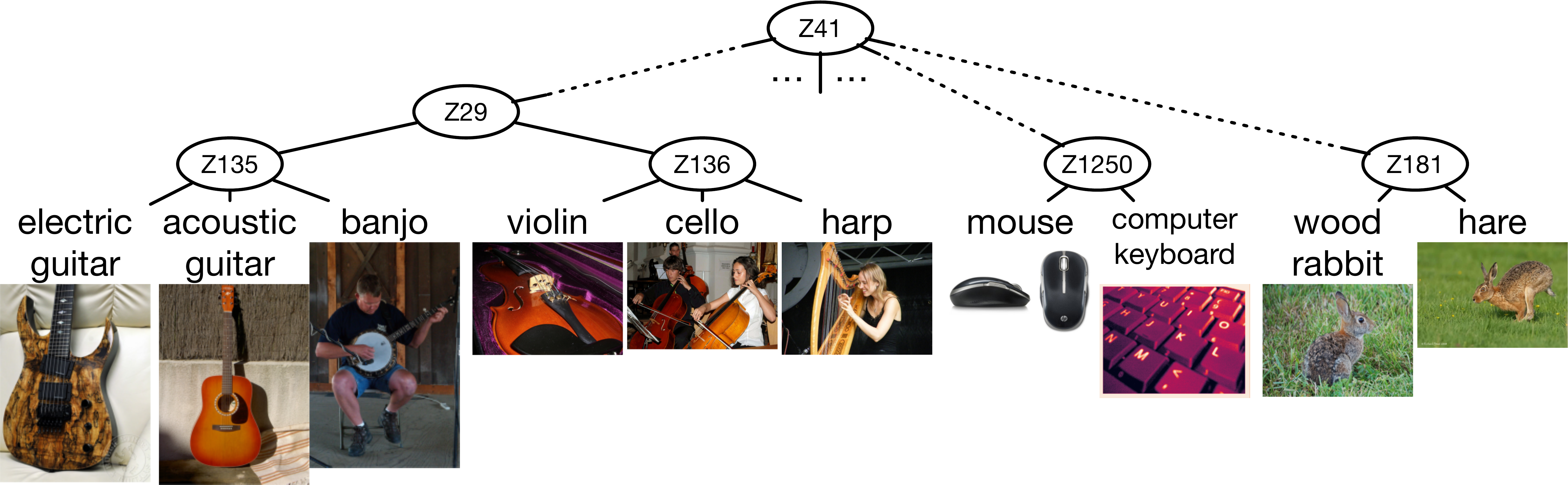}
	\caption{ A part of the latent tree model built from the outputs of ResNet50 on ImageNet training examples.  Solid lines are direct connections, and dashed lines are indirect connections with intermediate nodes removed. Images  of the classes are displayed for visual reference.  They are not part of the model. The tree reveals co-occurrence patterns of class labels in classification outputs.
	}
	\label{hlta}
\end{figure*}

There are many methods that can be used for word clustering.  
We choose to use hierarchical latent tree analysis (HLTA) \citep{chen2017latent,zhang2016latent}
because it is developed specifically  to model word co-occurrence in documents. 
We leave it to future work to evaluate other methods for this step.

HLTA is based on hierarchical latent tree models (HLTM), which
  are Bayesian networks with multiple levels of latent variables. An example is shown in   Fig.~\ref{hlta}.  The idea is to model correlations among 
  observed variables (the leaf nodes) using a tree of latent variables. Given a dataset on the observed variables, HLTA aims to find the model that maximizes the Bayes Information Criterion \citep{schwarz1978estimating}.

We  performed HLTA  on a collection of short documents obtained using ResNet50 on the training examples of ImageNet.
Fig.~\ref{hlta}  shows a part of the structure of the resulting model. \footnote{The entire model structure is in our GitHub repository.} 
The variables at the bottom level, level 0, are  binary variables that represent the presence/absence of words in a document. The latent variables at level 1 are introduced during the analysis to model word co-occurrence patterns, e.g.,  $Z181$ for the  co-occurrence of  {\tt hare} and {\tt wood-rabbit}, and  $Z1250$ for the  co-occurrence of  {\tt mouse} and {\tt computer-keyboard}.
Latent variables at level 2 are introduced during the analysis to model the co-occurrence of the patterns at level 1, e.g., $Z29$ for the co-occurrence of the patterns $Z135$ and $Z136$.

Each node in the tree  defines a cluster of class labels, which consists of the labels in the subtree rooted at the node. Some of the clusters given by level-1 nodes, for instance \{{\tt hare}, {\tt wood-rabbit}\} and
\{{\tt electric-guitar}, {\tt accoustic-guitar}, {\tt banjo}\}, consist of
visually similar classes that are difficult for the classifier to discriminate.  They are often competing labels for the same object/region in the input image, and hence all appear as top classes in classification output.

Class labels in some other clusters are not visually similar. One example is  \{{\tt mouse}, {\tt computer-keyboard}\}.  The two classes are grouped together nonetheless because mouses and keyboards tend to co-occur in  images, and hence co-occur as top classes in classification output.  Due to the co-occurrence, classifiers often have trouble in deciding which of them to use to label an image.
If we think of a {\em composite object} {\tt mouse+computer-keyboard}, then this co-occurrence cluster is no different from the visually similar clusters above: Different labels in the cluster are competing labels for the same (composite) object. See Fig.~\ref{composite} for example of explanations on the composite object.

There is a hierarchy behind the ImageNet classes that was derived from WordNet \citep{miller1995wordnet}.
While there are some similarities, our latent tree differs from
the WordNet hierarchy significantly. For example,  {\tt screwdriver} and {\tt syringe} are
far apart in WordNet, but close to each other in our latent tree due to their visual similarity. See Fig.~\ref{fig.syrdinge.intro}.

\section{Creating Contrastive Explanations}
\label{sec.cwox}

Suppose we want to explain the behaviors of a classification model $m$.
In our approach, the first step is to build a latent tree $T$ for all the class labels as described in the previous section. This is done in an {\em offline phase}.

During the {\em online phase}, we create explanations for the outputs of $m$ on individual inputs. For each input image $\bx$, we feed it to $m$ to get
the  $K$ top classes in the output. The value of $K$ can either be a predetermined number (e.g.,5), or the number of top classes  whose total probability exceeds a threshold (e.g., 0.95).

To divide the top classes into confusion clusters, we first restrict the latent tree $T$ from the offline phase onto those classes to obtain a subtree, and then cut the subtree at level 1 to get clusters of labels.  
 In our running example in Fig.~\ref{c-wox}, there are 5 top classes:
{\tt cello}, {\tt violin}, {\tt acoustic-guitar}, {\tt banjo}   and {\tt electric-guitar}.  By restricting the latent tree in Fig.~\ref{hlta}
onto those classes and cutting the resulting subtree at level 1, we get the following two clusters: {\tt \{cello,  violin\}}  and {\tt \{acoustic-guitar,  banjo, electric-guitar\}}.

In general, suppose the $K$ top classes are divided into $I$ confusion clusters $\bC=\{\bC_1$, \ldots, $\bC_I\}$, and  each cluster $\bC_i$ consists of $J_i$ class labels $\bC_i=\{c_{i1}$, \ldots, $c_{iJ_i}\}$.  To explain the top classes, CWOX-2s generates a collection of {\em contrastive heatmaps} in two stages:
\begin{enumerate}
\item For each confusion cluster $\bC_i$, create a heatmap to highlight the pixels that support $\bC_i$ over other clusters; 
\item In each  $\bC_i$, create a heatmap for each class $c_{ij}$ to highlight the pixels that support 
$c_{ij}$  over other classes. 
\end{enumerate}

\subsection{Base Explainers}
In CWOX-2s, contrastive heatmaps are created from saliency maps for individual classes. 
A {\em saliency map} for a classes $c$ 
aims to highlight the pixels that are, according to the model $m$, important for the class. The more important a pixel is to the class, the higher its saliency value.  It is usually computed from either the probability $P_m(c|\bx)$ or the logit $z_c(\bx)$ of the class.  
Saliency maps can be generated by a variety of IOX methods, including backpropagation-based techniques such as Guided Backpropagation \citep{springenberg2014striving}, DeepLIFT  \citep{shrikumar2017learning}, Grad-CAM \citep{selvaraju2017grad}; forward propagation-based techniques like RISE \citep{petsiuk2018rise}, and local approximation methods like LIME \citep{ribeiro2016should}.  They will be referred to as {\em base explainers}
in the context of CWOX-2s.

The concept of saliency map can easily be generalized to clusters of classes.
A cluster $\bC$ of classes can be viewed as a {\em compound class} with
probability and logit given as follows:
\[P_m(\bC|\bx) = \sum_{c \in \bC}P_m(c|\bx),
\hspace{0.3cm} z_{\bC}(\bx) = \log \sum_{c \in \bC} e^{z_c(\bx)}.\]
Saliency maps can be generated for the cluster $\bC$ in the same way as for individual classes. 

\subsection{Contrastive Heatmaps}

Let $H_{\bC_i}$ and $H_{\bC \backslash \bC_i}$ be
 saliency maps for a confusion cluster $\bC_i$ and the union of all other confusion clusters respectively.  $H_{\bC_i}$ and
$H_{\bC \backslash \bC_i}$ presumably highlight the pixels that are, according to the model $m$, important for 
 $\bC_i$ and $\bC \backslash \bC_i$ respectively. For a given pixel 
$x$, the difference $H_{\bC_i}(x) - H_{\bC\backslash {\bC_i}}(x)$ measures
the importance of $x$ to $\bC_i$ relative to $\bC\backslash {\bC_i}$. Consequently, we 
use the following heatmap to contrast $\bC_i$ against other confusion clusters:
\begin{eqnarray}
	&	\hat{H}_{\bC_i}=& \left\{ \begin{array}{ll}
		ReLU[H_{\bC_i} - H_{\bC\backslash {\bC_i}}]& \mbox{if } I>1;\\
		{H}_{\bC_i}& \mbox{if } I=1,
	\end{array} \right. 
\end{eqnarray}
Note that  ReLU is used so as to focus on the evidence for cluster $\bC_i$
rather than that against it.

Next, consider all the classes in a confusion cluster $\bC_i$.  
Let $H_{c_{ij}}$ and $H_{\bC_i \backslash c_{ij}}$ be
saliency maps for a class $ c_{ij} \in \bC_i$ and all other classes in the same cluster respectively. We 
use the following heatmap to contrast $c_{ij}$ against the other classes:
\begin{eqnarray}
	&	\hat{H}_{c_{ij}}=&  \left\{ \begin{array}{l} \supp(\hat{H}_{\bC_i}) \times ReLU[{H}_{c_{ij}} -H_{\bC_i\backslash c_{ij}}]   \\ 
		\mbox{if } J_i>1;\\
		\supp(\hat{H}_{\bC_i}, \epsilon) \times {H}_{c_{ij}}  \\
		\mbox{if } J_i=1.
	\end{array}  \right. 
\end{eqnarray}
Note that the contrastive heatmap for $c_{ij}$ is restricted to $\supp(\hat{H}_{\bC_i})$.
This means that, when identifying contrastive
evidence for classes in the $\bC_i$, we focus only on the evidence supportive of the cluster. 

An overall description of CWOX-2s is given in Algorithm \ref{alg:algorithm}.
As alluded to earlier, a confusion cluster $\bC_i$ consists of classes that are competing labels for the same region in the input image. The first step of CWOX-2s aims to highlight that region, and hence is about object localization. The second step  of CWOX-2s aims to pinpoint at the evidence for each of the competing classes. It is about class discrimination. Class discrimination requires more fine-grained information.  {Some base explainers can facilitate this desiderata. For instance in  Grad-CAM, one needs to specify a pivot layer where multiple feature maps are aggregated into one heatmap using gradients from the output layer. The further away the pivot layer is from the output layer, the more fine-grained is the heatmap. In RISE, one needs to specify a mask size and a pixel mask probability.  The smaller the mask size and the pixel mask probability, the more fine-grained the resulting heatmap.}

The idea of subtracting two saliency maps to create a contrastive heatmap was first proposed by  \citet{shrikumar2017learning,zhang2018top}. Alternatively, one can multiply one saliency map with the ``inverse" of the other \citep{wang2020scout}.  

\begin{algorithm}[tb]
	\caption{CWOX-2s}
	\label{alg:algorithm} 
	{I. OFFLINE PHASE}\\
	\textbf{Input}: A classification model $m$; a dataset $S$.\\
	\textbf{Do}:
	\begin{algorithmic}[1] 
		\STATE Feed each example $\bx$ in $S$ to $m$ to get a list (document) of top class labels.
		\STATE Run HLTA  on the documents to get a latent tree $T$.
	\end{algorithmic} 	

	{II. ONLINE PHASE}\\
	\textbf{Input}: A test example $\bx$; a base explainer.\\
	\textbf{Do}:
	\begin{algorithmic}[1] 
		\STATE Feed $\bx$ to $m$ to get a list of top class labels.
		\STATE Restrict $T$ to those labels to get a subtree
		\STATE Partition the labels into confusion clusters by cutting the subtree at level 1.
		\STATE Create a heatmap to contrast each confusion cluster against other clusters using Equation (1).
		\STATE	In each cluster, create  a heatmap to contrast each class in the cluster
		against other classes using Equation (2).
	\end{algorithmic} 	

\end{algorithm}

\begin{figure*}[t]
	\centering
	\begin{tabular}{ccc}
		\includegraphics[width=5.15cm]{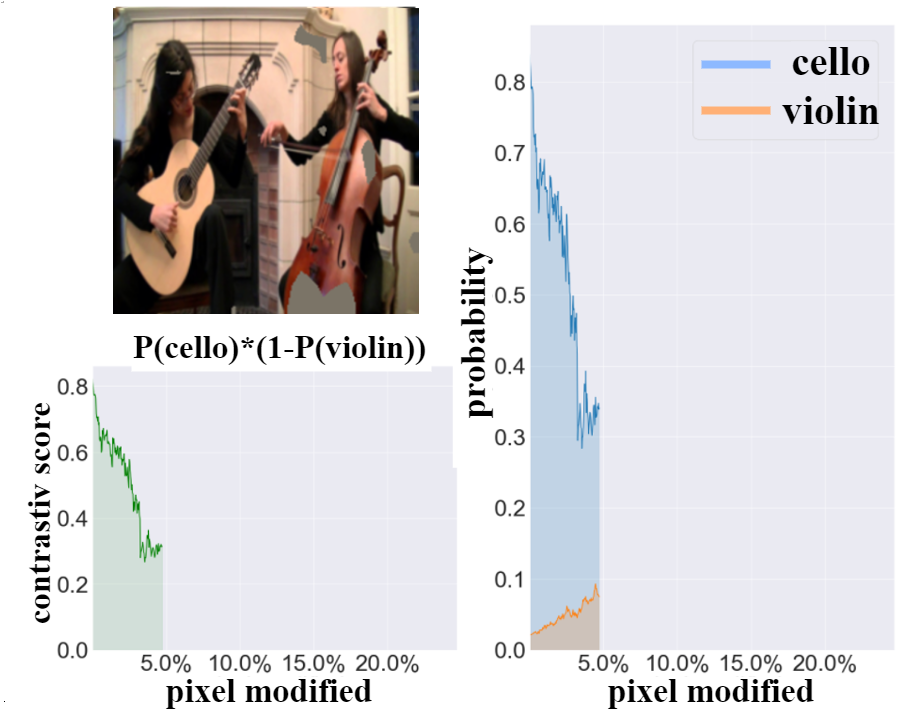} &
		\includegraphics[width=5.15cm]{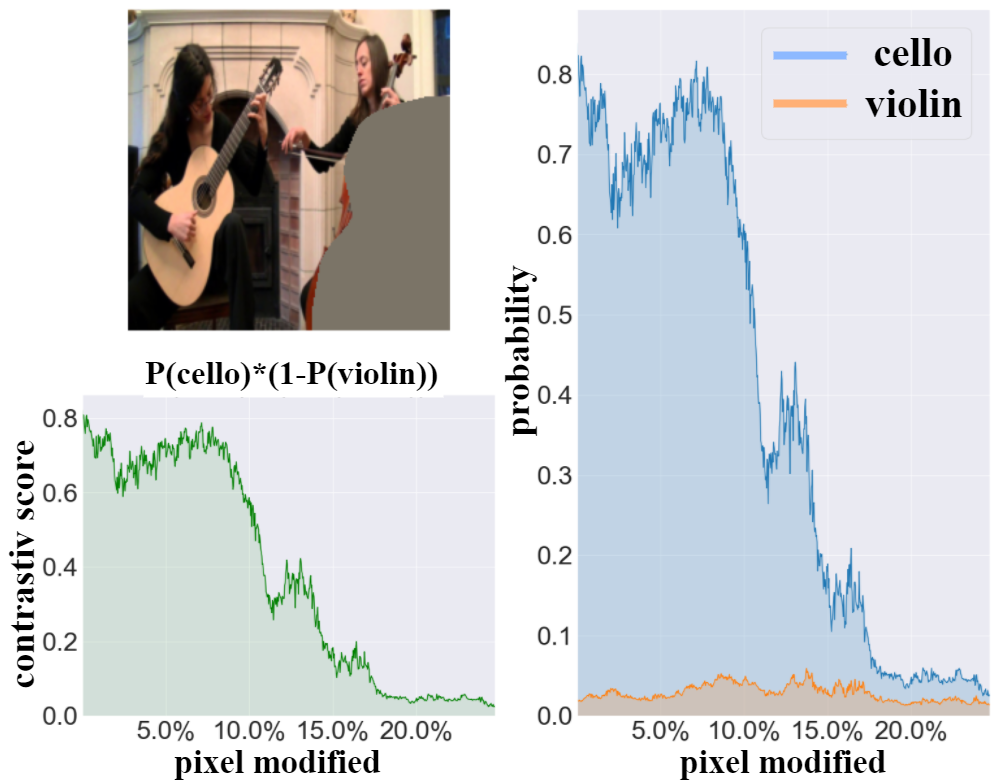} &
		\includegraphics[width=5.15cm]{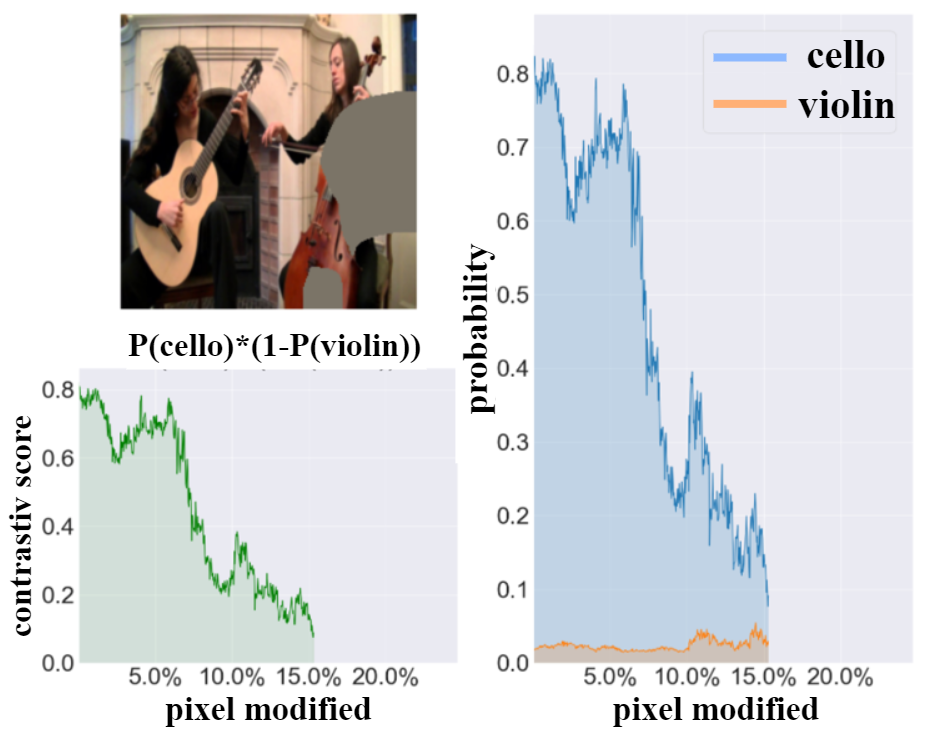}
		 \\
		
		{  (i) CWOX-2s: $n_{\delta}=2,377$, }    &  { (ii) SWOX:  $n_{\delta}=12,402 $,} &  { (iii) CWOX-1sA:  $n_{\delta}=7,686 $,} \\
		CAUC = 0.023, CDROP = 0.505; &  CAUC = 0.033, CDROP = 0.309; &CAUC = 0.033, CDROP = 0.364.
	\end{tabular}
	
	\caption{ Changes in the probabilities $P({\tt cello})$ and $P({\tt violin})$ and the contrastive score $P({\tt cello}) \times (1-P({\tt violin}))$ as $\delta$-salient pixels are deleted according to the order induced by: (i) the CWOX-2s heatmap in Fig.~\ref{c-wox} (b.1); (ii) the SWOX heatmap in  Fig.~\ref{fig.cello.intro} (b.1); and (iii) the CWOX-1sA heatmap in Fig.~\ref{fig.cello.intro} (c.1).} 
		
	\label{fig.cf}
\end{figure*}

\begin{table*}[t]
	\centering
		\caption{ Average CAUC scores on the  ImageNet examples (\textbf{smaller $\downarrow$} CAUC  indicates better contrastive faithfulness).} 
	\begin{tabular}{c|cc|cc} \hline
		&  \multicolumn{2}{c}{ResNet50} &  \multicolumn{2}{|c}{GoogleNet} \\ \hline
		& { Grad-CAM}   & { RISE}             &   { Grad-CAM}  &  { RISE} \\ \hline
		{ SWOX} &  { $7.54 \times 10^{-3}$}   & { $5.18 \times 10^{-3}$}  & { $5.93 \times 10^{-3}$}
		& { $3.36 \times 10^{-3}$} \\
		{ {CWOX-1sA}}   & { {$7.19\times 10^{-3}$}}& {{$4.65\times 10^{-3}$}}& {{$5.37\times 10^{-3}$}}& {{$3.12\times 10^{-3}$}} \\
		{ {CWOX-1sB}}   & {{$7.68\times 10^{-3}$}}& {{$4.96\times 10^{-3}$}}& {{$6.12\times 10^{-3}$}}& {{$3.24\times 10^{-3}$}} \\
		{ CWOX-2s} & {${\bf 5.78 \times 10^{-3} }$}  & {${\bf 4.08 \times 10^{-3}}$}  & {${\bf 4.47 \times 10^{-3}}$}        & {${\bf 2.78 \times 10^{-3}}$} \\ \hline
	\end{tabular}
	\label{table.cauc}

\	

	\centering
		\caption{ Average CDROP scores on the ImageNet examples (\textbf{larger $\uparrow$} CADROP  indicates better contrastive faithfulness).}
	\begin{tabular}{c|cc|cc}
		\hline
		& \multicolumn{2}{c|}{ResNet50} & \multicolumn{2}{c}{GoogleNet} \\ \hline
		& { Grad-CAM} & { RISE} & { Grad-CAM}  & { RISE} \\ \hline
	{	SWOX}      &           { $6.84\times 10^{-2}$}                                   & {$8.19\times 10^{-2}$}                      &         { $6.56\times 10^{-2}$}                                 & {$7.68\times 10^{-2}$}                  \\
		{ {CWOX-1sA}}   & {{$7.01\times 10^{-2}$}}& {{$8.35\times 10^{-2}$}}& {{$6.59\times 10^{-2}$}}& {{$7.73\times 10^{-2}$}} \\
		{ {CWOX-1sB}}   & {{$5.22\times 10^{-2}$}}& {{$7.01\times 10^{-2}$}}& {{$5.05\times 10^{-2}$}}& {{$6.32\times 10^{-2}$}} \\
		{ CWOX-2s}            &          {${\bf 8.21\times 10^{-2}}$}                               & {${\bf 8.97\times 10^{-2}}$}                       &               {${\bf 7.64\times 10^{-2}}$}                              & {${\bf 8.32\times 10^{-2}}$}                     \\
		\hline
		
	\end{tabular}

	\label{table.cdrop}
	
	\

	\centering
		\caption{ Performances of CWOX-2s with four base explainers. Here, $\bar{n}_{\delta}$ stands for average number of $\delta$-salient pixels.}
	\begin{tabular}{c|ccc|ccc}
		\hline
		& \multicolumn{3}{c|}{ResNet50} & \multicolumn{3}{c}{GoogleNet} \\ \hline
		{ 	Base Explainer} & { $\bar{n}_{\delta}$} &  { CAUC $\downarrow$} & { CDROP $\uparrow$} & {  $\bar{n}_{\delta} $  } &  { CAUC $\downarrow$} &{ CDROP $\uparrow$} \\ \hline
		{ Grad-CAM  }     &         {  2,029 }     &  { $3.11\times 10^{-3}$    }                    &  { $8.01\times 10^{-2}$  }                    &        {  2,181   }      & { $1.80\times 10^{-3}$  }                       & { $7.46\times 10^{-2}$    }              \\
		{	MWP     }       &       {   4,194}   & { $3.12\times 10^{-3}$ }                     &  { $7.37\times 10^{-2}$ }                     &              { 3,026}
		& { $1.77\times 10^{-3}$    }                    & { $5.85\times 10^{-2}$ }                    \\
		{	LIME    }       &      {   2,464  }   &{  $3.40\times 10^{-3}$    }                         & { $4.89\times 10^{-2}$  }                       &               { 2,351}  & { $1.92\times 10^{-3}$  }                     & { $3.64\times 10^{-2}$   }                   \\
		{ RISE   }        &      {      1,282 }   & { {${\bf 3.07\times 10^{-3}}$ }}                      & { ${\bf 8.97\times 10^{-2}}$ }                     &           {   1,105  }  & { ${\bf 1.72\times 10^{-3}}$   }                       &  { ${\bf 8.32\times 10^{-2}}$ }                     \\ \hline
		
	\end{tabular}

	\label{b.table1}

\end{table*}

\section{Empirical Evaluations}

In this section, we evaluate CWOX-2s against several other WOX methods to explain all top classes. The evaluations are in terms of  
the faithfulness and interpretability of the explanations.  Here,
{\em faithfulness} refers to an explanation's ability to accurately reflect the function learned by the model \citep{selvaraju2017grad,petsiuk2018rise}, while {\em interpretability} refers to its ability to provide a clear understanding of the relationship between input and output for human users \citep{ribeiro2016should,doshi2017towards}.

We have presented three methods, SWOX, CWOX-1s and CWOX-2s, for explaining all top classes. CWOX-1s has two possible variants. CWOX-1sA obtains
a heatmap for each class by subtracting saliency maps, i.e.,  $ReLU[H_c - H_{\bC\backslash c}]$,  similar to the contrastive heatmaps created in CWOX-2s. On the other hand, CWOX-1sB multiplies 
$H_c$ with the ``inverse" of $H_{\bC\backslash c}$. {The second variant was proposed earlier in \citep{wang2020scout}}, where it is called {\em SCOUT}.
As will be seen, CWOX-1sA  significantly 
outperforms the CWOX-1sB.  Hence, we do not consider the B-variant of CWOX-2s.

To evaluate CWOX-2s and the  three baselines, we use them to explain the outputs  of  GoogleNet \citep{szegedy2015going} and ResNet50 \citep{he2016deep} on a subset of randomly selected 10,000 images from the ImageNet validation set \citep{deng2009imagenet}. For each image, we apply the WOX methods to explain its top $K$ predicted classes with $K=\min \{5, Cum(0.95)\}$, where $Cum(0.95)$ is a function to return the smallest number of top classes with a cumulative probability greater than 0.95. Two base explainers, namely Grad-CAM \citep{selvaraju2017grad} and RISE \citep{petsiuk2018rise}, are used in the experiments.

\subsection{Faithfulness to Model}

{\bf Rationale for Evaluation Metrics}: An IOX method aims to reveal the evidence a model relies on to predict a particular class. IOX explanations (saliency maps) are often evaluated in terms of
their {\em faithfulness} to a model. Ideally, a faithful saliency map should highlight important pixels for the class, and removing pixels with high saliency values  should decrease the class probability. This concept gives rise to a widely-used metric, the {\em deletion AUC} metric \citep{samek2016evaluating,petsiuk2018rise}.

Different from IOX, CWOX-2s aims to reveal the evidence that a model uses to discriminate between classes.  Consequently, CWOX-2s explanations should be evaluated
in terms of their {\em contrastive faithfulness} to a model $m$, i.e.,  how effective they are at revealing the evidence that the model relies on to discriminate between different classes.


Suppose that a model $m$ has reasons to believe that an input $\x$ belongs to a class $c$, but cannot rule out the possibility of it belonging to some other classes $\bC'$. If a heatmap $H$ is contrastively faithful to $m$, then it should give high values to the pixels that  $m$
considers strongly supportive of $c$ relative to $\bC'$. The deletion of such high-value pixels should lead to fast decrease in the probability of $c$ and an increase in that of $\bC'$. {To be more specific, let there be totally $n$ pixels, and $x_1, \ldots, x_{n}$ be an enumeration of pixels in descending order of $H(x)$. Let $\x_{[r, n]}$ be the resulting image of deleting  the first $r-1$ pixels from the input image $\x$.  If  $H$ is contrastively faithful to $m$, then the probability $P_m(c|\x_{[r, n]})$ would decrease quickly with $r$ and $P_m(\bC'|\x_{[r, n]})$ would increase with it. Thus, the {\em contrastive score} defined below would decrease quickly: 
\begin{eqnarray}
s(r) =  P_m(c|\x_{[r, n]}) (1- P_m(\bC'|\x_{[r, n]})).
\label{contrast_score}
\end{eqnarray}
As an example, consider the CWOX-2s heatmap (b.1) in Fig.~\ref{c-wox}.
It presumably reveals the evidence that ResNet50 considers supportive of ${\tt cello}$ relative  to ${\tt violin}$.  Fig.~\ref{fig.cf} (i) shows what happens
when pixels are deleted from the input image
according to order induced by the CWOX-2s heatmap.  We see that  $P({\tt cello})$ decreases and $P({\tt violin})$ increases. Consequently, the contrastive score $P({\tt cello}) \times (1-P({\tt violin}))$ decreases.
Fig.~\ref{fig.cf} (ii) shows what happens
when pixels are deleted according to the order induced by SWOX heatmp shown in  Fig.~\ref{fig.cello.intro} (b.1).  We see that, compared to the CWOX-2s heatmap, the contrastive score of the SWOX heatmap drops more slowly at the beginning. Although a bigger drop  is achieved later,  it is at the expense of deleting many more pixels. Those indicate that the SWOX heatmap is less effective than the CWOX-2s at pinpointing at the evidence that ResNet50 relies  on to discriminate {\tt cello} from {\tt violin}. Additionally, Fig.\ref{fig.cf} (iii) shows the results when pixels are removed based on the order suggested by the CWOX-1sA heatmap shown in Fig.\ref{fig.cello.intro} (c.1). Although it has a smaller $n_\delta$ and a larger CDROP compared to SWOX, its overall performance is still notably inferior to the CWOX-2s results that are shown in (i).

{\bf Evaluation Metrics}: We propose two quantitative metrics to evaluate the
contrastive faithfulness of a heatmap to the target model.  The first one is the area under  the contrastive score curve, or {\em contrastive AUC (CAUC)} for short:
\begin{eqnarray}
	\label{eq.cauc}
	CAUC(H, m|\x, c, \bC') &=&  \frac{1}{n}\sum_{r=1}^{n_{\delta}}s(r),
\end{eqnarray}
where $s(r)$ is the contrastive score  defined in Equation (\ref{contrast_score}) and $n_{\delta}$ is the number of 
{\em $\delta$-salient pixels}. A pixel $x$ is considered $\delta$-salient if its saliency value, denoted by $H(x)$, is greater than or equal to $\delta$ times the maximum saliency value across all pixels in the heatmap $H$. In other words, such pixels have a saliency value of $H(x) \geq \delta\max_{x' \in \x} H(x')$. The rationale behind the $\delta$-salient pixels is twofold: First, it enables the evaluation to concentrate on the most salient pixels, and second, it helps to exclude the numerous zero-valued pixels that are frequently presented in CWOX-2s heatmaps and lack any meaningful ordering. In our experiments, $\delta$ is set to 0.5.

Similar to the deletion AUC, smaller CAUC scores indicate better heatmaps in terms of contrastive faithfulness.  
In Fig. \ref{fig.cf}, the curves are shown only for the first $n_{\delta}$  pixels. Different heatmaps may have different $n_{\delta}$. CAUC scores computed over different numbers of pixels are not comparable. Consequently, when comparing two or more heatmaps, we use the minimum of numbers of salient pixels to calculate the CAUC scores for all of the heatmaps.

One drawback of the CAUC metric is that, when comparing two heatmaps, it does not consider all the $\delta$-salient pixels in some of the heatmaps. To take all the $\delta$-salient pixels into consideration, we propose another metric called the {\em weighted drop in contrastive score (CDROP)}:
\begin{eqnarray}
	\label{eq.cdrop}
	CDROP(H, m|\x, c, \bC') = \frac{s(1) - s(n_{\delta}+1)}{\log_2 (1+ \frac{\max\{n_{\delta}, \tau\}}{\tau})}\ ,
\end{eqnarray}
\noindent where $\tau$ is a hyperparameter. The score is a combination of two factors. The first factor $s(1) - s(n_{\delta}+1)$ is the drop in the contrastive score
due to the deletion of all the salient pixels. The second factor
$\log_2 (1+ \frac{\max\{n_{\delta}, \tau\}}{\tau})$ is
a logarithmic penalty factor for  $n_{\delta}$ when it exceeds $\tau$,
which is set at $0.05 n$ in our experiments (i.e., 5\% of the total number of pixels). It captures the intuition that too many salient pixels can be distracting to a human user, and
is motivated by the Weber-Fechner law. This Weber-Fechner law posits that the intensity of human sensation grows in proportion to the {\em logarithm} of an increase in energy, rather than increasing at the same rate as the energy. Larger CDROP scores indicate better heatmaps in terms of contrastive faithfulness.


{\bf Results}:  Tab. \ref{table.cauc} presents the CAUC scores of the four methods. These scores are averaged across all 10,000 test images, and for each test image, all pairs $(c_{ij}, \bC_i \backslash {c_{ij}})$ are considered. These pairs are composed of a top class $c_{ij}$ and all other classes $\bC_i \backslash {c_{ij}}$ from the same confusion cluster $\bC_i$. Tab. \ref{table.cdrop} show the corresponding CDROP scores. We see that the CAUC scores of CWOX-2s are significantly lower than those of the baselines, and its CRDOP scores are significantly higher. Those indicate that  the explanations produced by CWOX-2s are more contrastively faithful to the models than the baselines.  It is also interesting to note that CWOX-1sA is inferior to CWOX-1sB in all cases, and sometimes it is even inferior to SWOX.

Apart from comparing various WOX methods, Tab. \ref{b.table1} presents a performance comparison of CWOX-2s using four different IOX methods as base explainers, including two backpropagation-based methods, Grad-CAM \citep{selvaraju2017grad} and MWP \citep{zhang2018top}, as well as two forward propagation-based methods, LIME \citep{ribeiro2016should} and RISE \citep{petsiuk2018rise}. We can see that RISE outperforms others in contrastive faithfulness metrics, with the lowest CAUC score and the highest CDROP score. It also has the fewest salient pixels, which indicates the precision in identifying crucial evidence. Grad-CAM follows, then MWP, which produces numerous salient pixels and is less precise. LIME has the weakest performance among the four. Appendix A details the setup for each base explainer in CWOX-2s and provides examples for comparing the contrastive faithfulness across them.

\begin{figure}[t]
	\centering
\includegraphics[height=6.64cm]{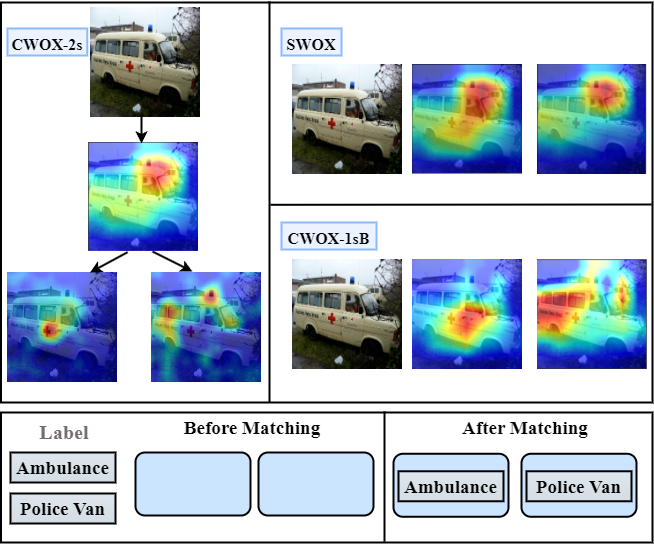} 
	\caption{ In the user study,
		heatmaps for pairs of confusing labels are displayed.  A user is asked to match the labels with the heatmaps.}
	\label{fig.hs}
\end{figure}

\subsection{Interpretability to Users}

How well does a WOX method help human users understand the evidence that a model relies on to discriminate between classes? To answer this question, we have conducted an user study following the {\em forward simulation} protocol \citep{ribeiro2016should,doshi2017towards,nunes2017systematic,lage2019evaluation,tjoa2020survey}. 
As shown in Fig.~\ref{fig.hs}, we display the heatmaps for pairs of
confusing labels alongside the input image, and ask  users to match the heatmaps with the labels.
A correct matching would indicate that a user understands what pixels the model considers important for each of the two labels.

The study was conducted on the predictions by ResNet50 on a collection of images from the ImageNet validation set.  For each image, a pair of confusing top classes was selected based on the latent tree $T$ from the offline phase.  CWOX-2s, CWOX-1sB and SWOX were included in the study. To make the study manageable, we did not consider all possible combinations 
of image classification models, WOX methods, and base explainers.  We also limited the choices of input images and confusing class labels. See {Appendix B} for the details.  CWOX-1sB  was chosen over CWOX-1sA because it is based on previous work \citep{wang2020scout}, and also because CWOX-1sA would simplify to CWOX-2s when there is only one confusion cluster for the inputs.  RISE was used as the base explainer due to its proven superior contrastive faithfulness compared to other base explainers (as shown in Tab. \ref{b.table1}).

\begin{table}[t]
	\centering
	\caption{ Results of the user study in the \underline{expert group} ($\pm$ 95\% confidence interval).}
\begin{tabular}{c|ccc} \hline
	&	SWOX &  CWOX-1sB   & CWOX-2s         \\ \hline
	{\smaller Accuracy} & {0.45$\pm$0.048} &{0.57$\pm$0.088} & {\bf 0.83}{$\pm$0.092} \\ \hline
	{\smaller Confidence} &{1.60$\pm$0.241} &{2.60$\pm$0.241} & {\bf 3.60}{$\pm$0.237}\\ \hline
\end{tabular}
\label{result_hs_e}
		
		\

\caption{ {Results of the user study in the \underline{non-expert group} ($\pm$ 95\% confidence interval).}}
\begin{tabular}{c|ccc} \hline
	&	SWOX &  CWOX-1sB   & CWOX-2s         \\ \hline
	{\smaller Accuracy} & {0.40$\pm$0.075} &{0.51$\pm$0.102} & {\bf 0.75}{$\pm$0.119} \\ \hline
	{\smaller Confidence} &{1.40$\pm$0.108} &{2.80$\pm$0.172} & {\bf 3.40}{$\pm$0.163}\\ \hline
\end{tabular}

\label{result_hs_n}

\end{table}

The user study consisted of two groups of participants.  The first group, referred to as the expert group, included postgraduate students enrolled in a machine learning course. These students had hands-on experience with deep computer vision models, such as training a CNN model. In contrast, the second group, known as the non-expert group, consisted of first-year undergraduate students who had no prior experience or knowledge in training deep learning models. Both groups had an equal number of participants, with 60 individuals in each group.

Each group was randomly divided into three subgroups, and  each subgroup
was responsible for only {\em one} of the three WOX methods.
As in previous XAI user studies
\citep{doshi2017towards,nguyen2018comparing,hase2020evaluating,fel2021cannot}, the participants went through a training phase before handling explanations for {\em new unseen examples}. Besides the matching task, they were also asked to rate their confidence in their answers on a scale of 1 to 5, with 1 meaning  ``not sure'' and 5 meaning ``completely sure''.

The results are shown in Tab. \ref{result_hs_e} and \ref{result_hs_n}. In both groups, the accuracy and confidence of CWOX-2s were significantly better than CWOX-1sB and SWOX. These results indicate that CWOX-2s is more effective in helping both novice and expert users understand the evidence used by the model to discriminate between classes. Interestingly, compared to the non-expert group, the expert users showed higher accuracy with narrower confidence intervals, especially among those responsible for CWOX-2s. This suggests that expert users can acquire more information about model behaviors from CWOX-2s explanations than non-expert users.

To get a concrete feeling about the superiority of CWOX-2s, imagine completing the matching tasks shown in Fig. \ref{fig.hs}.
Among the two second-stage  CWOX-2s heatmaps  (those at the bottom),
the one on the left highlights the red cross, while the one on the right does not. Hence, the former should be obviously matched with {\tt ambulance} and the latter with {\tt police van}. The matching task is relatively more challenging with heatmaps by SWOX and CWOX-1sB.

\subsection{Visual Examples}

In this section, we provide two more visual examples with Grad-CAM as the base explainer. More examples including examples with different base explainers are in {Appendix C}.

The first example (Fig. \ref{composite}) illustrates the differences between SWOX and CWOX-2s when used to explain the output of an 
 an image that contains both keyboard and mouse.
 As discussed in Section \ref{sec.hlta},  {\tt mouse}
and {\tt computer keyboard} are grouped together in the latent tree as they often co-occur in images. Consequently, CWOX-2s first identifies evidence for the composite object {\tt mouse+computer-keyboard} (c),
 and then the evidence for {\tt keyboard} against {\tt mouse} (c.1) and the evidence for {\tt mouse} against {\tt keyboard} (c.2).
 The CWOX-2s explanations are clear and discriminative.  However, the SWOX 
 explanations for  {\tt mouse}
 and {\tt computer keyboard} are very similar to each other, and hence are not discriminative.

The second example (Fig. \ref{guitars}) shows the difference between CWOX-1sA and CWOX-2s when explaining the output of an image that include the {\tt electric guitar} and {\tt acoustic guitar} among the top predicted classes. While the CWOX-2s explanations (b.1-2) provide discriminative information for the two visually similar class, CWOX-1sA explanations (d.1-2) exhibit noticeable overlap in highlighted regions, making it difficult to understand what features the model relies on to distinguish the two classes.

\begin{figure}[t]
	\centering
		 \includegraphics[height=5.2cm]{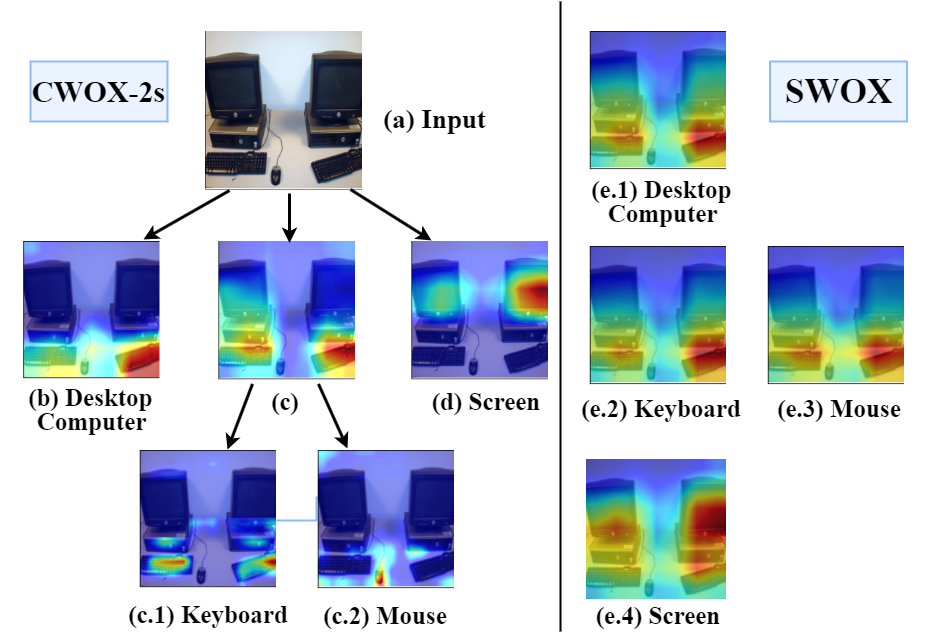}\\
	\caption{ SWOX and CWOX-2s explanations of the output of ResNet50
		on an image that contains keyboard and mouse.}
	\label{composite}

		\centering
	\includegraphics[height=5.2cm]{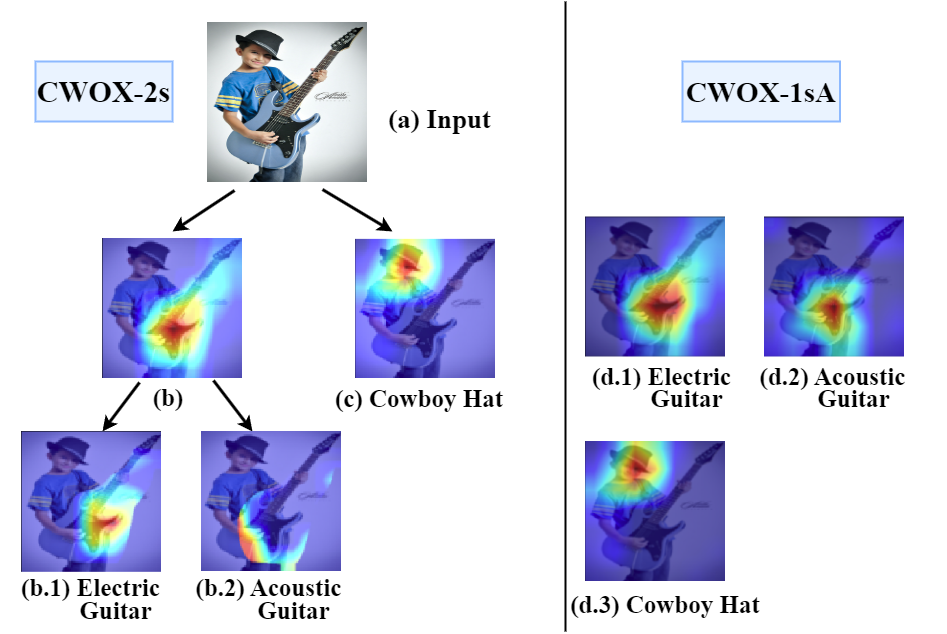}\\
	\caption{ CWOX-2s and CWOX-1sA explanations of the output of ResNet50 on an image with {\tt electric guitar} and {\tt acoustic guitar} among the top classes.}
\label{guitars}
\end{figure}

\section{Conclusion}

We propose a novel post-hoc local explanation method called CWOX-2s for image classification.  Unlike most previous methods,  CWOX-2s explains all top classes in the output rather than one individual class.  The key technical contribution is a principled method for determining how to contrast the top classes against each other. Recently,  a new conceptual framework for XAI termed  {\em evaluative AI} is proposed \citep{miller2023explainable}, which stresses the use of
XAI to ``provide evidence for and against decisions made by people,
rather than provide recommendations to accept or reject".  CWOX-2s aligns with this framework nicely.
Empirical results show that, in comparison with alternative methods that explain all top classes, CWOX-2s produces explanations that are more faithful to the model and more interpretable to human users. Furthermore, we propose two metrics for evaluating contrastive explanations, namely Contrastive AUC (CAUC) and Weighted Drop in Contrastive Score (CDROP).

\begin{acknowledgements}
We thank the deep learning computing framework MindSpore (\url{https://www.mindspore.cn}) and its team for the support on this work. Research on this paper was supported in part by Hong Kong Research Grants Council under grant 16204920. Weiyan Xie was supported in part by the Huawei PhD Fellowship Scheme. We thank Prof. Janet Hsiao, Yueyuan Zheng, Luyu Qiu and Yunpeng Wang for valuable suggestions and discussions. We thank April Hua Liu for organizing the user study with non-experts. 
\end{acknowledgements}

{
\bibliography{xie_130.bib}
}


\end{document}